\documentclass{article}

\usepackage{arxiv}

\usepackage[utf8]{inputenc} % allow utf-8 input
\usepackage[T1]{fontenc}    % use 8-bit T1 fonts
\usepackage{hyperref}       % hyperlinks
\usepackage{url}            % simple URL typesetting
\usepackage{booktabs}       % professional-quality tables
\usepackage{amsfonts}       % blackboard math symbols
\usepackage{nicefrac}       % compact symbols for 1/2, etc.
\usepackage{microtype}      % microtypography
\usepackage{lipsum}		% Can be removed after putting your text content
\usepackage{graphicx}
\usepackage{amsmath}
\usepackage{array}

\title{A Brief Summary of Interactions Between Meta-Learning and Self-Supervised Learning}

\date{} 					% Or removing it

\author{ \href{https://orcid.org/0000-0001-7431-1619}{\includegraphics[scale=0.06]{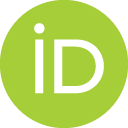}\hspace{1mm} Huimin Peng}\thanks{
		Thank you for all helpful comments! Feel free to leave a message about comments on this manuscript. In case I did not receive email, my personal email is \texttt{974630998@qq.com}. Thanks to github.com/kourgeorge/arxiv-style for this pdf latex template.} \\
	\texttt{peng.huimin.pennie@gmail.com} \\
}

\begin{document}
\maketitle

\begin{abstract}
This paper briefly reviews the connections between meta-learning and self-supervised learning. Meta-learning can be applied to improve model generalization capability and to construct general AI algorithms.
Self-supervised learning utilizes self-supervision from original data and extracts higher-level generalizable features through unsupervised pre-training or optimization of contrastive loss objectives. 
In self-supervised learning, data augmentation techniques are widely applied and data labels are not required since pseudo labels can be estimated from trained models on similar tasks. 
Meta-learning aims to adapt trained deep models to solve diverse tasks and to develop general AI algorithms. 
We review the associations of meta-learning with both generative and contrastive self-supervised learning models. 
Unlabeled data from multiple sources can be jointly considered even when data sources are vastly different. 
We show that an integration of meta-learning and self-supervised learning models can best contribute to the improvement of model generalization capability. 
Self-supervised learning guided by meta-learner and general meta-learning algorithms under self-supervision are both examples of possible combinations.
\end{abstract}

% keywords can be removed
\keywords{Meta-Learning \and Self-Supervised Learning \and Generative \and Contrastive \and Pretext task}

\section{Introduction}
	\label{introduction}

\subsection{Self-Supervised Learning}
\label{Self-SupervisedLearning}

Self-supervised learning utilizes self-supervision to train embedding functions (dimension reduction modules) that identify more generalizable and transferable high-level features. 
These features can be applied in downstream tasks to boost performance and pre-trained models can be used as initial models to avoid training from scratch. 

\subsubsection{Dimension Reduction}
\label{dimension-reduction}	

%dimension reduction in deep neural network
For high-dimensional input data, deep neural network performs dimension reduction and extracts meaningful features that are invariant to unimportant data transformations.
%from many to one, in classifier
In dimension reduction, a great proportion of information in the original data is discarded. 
For example, a deep neural network for image classification maps a high-dimensional input image to an image category scalar.
In the layer-wise structure of neural network, information loss at each layer is jointly optimized  through the backpropagation of loss gradients.
However, even so, dimension reduction has not reached a satisfying state, and the extraction of more generalizable and transferable features is still desired.
%noise, correlated features, that's why smoothing (convolutions) come into play
Pixels are usually noisy and correlated with each other in image data. 
%noise - convolutions and pooling
Convolution and pooling layers contribute to pixel-wise noise reduction.
%correlated features - centering and standardization
Centering and standardization modules reduce correlation between pixels.
Smoothing and convolution techniques reduce pixel-wise noise and correlation and play an important role in extracting critical features.
%correlation reduction techniques
Techniques regarding noise and correlation reduction should improve the generalization property of extracted features.

%for example, MNIST analysis, 
MNIST \cite{LecunYann1998} is a dataset containing images of hand-written digits.
%from image pixels to a class, from hundreds of data points to one class, coarsening data, averaging data, reduced pixels, still same answer, better immune to noise and correlation, most information is redundant
Every image includes 28*28 pixels, which are mapped to 1 scalar representing the image category. 
Neural network models solve MNIST classification problem with almost perfect prediction accuracy. Given MNIST images and a pre-trained neural network model, predicted labels are the same as true labels. We can see that true labels are now redundant information.
For hand-written digits images with higher resolution, coarsening or averaging image data reduces the number of pixels but still produces the same prediction accuracy. 
We can see that most pixels in the original high-fidelity image are also redundant information. 
Indeed, feature extraction modules should identify the optimal process of information loss to learn generalizable representations.  

%dimension reduction - from mega-high dimension to one
In some cases, dimension reduction from a high-dimension dataset to a scalar
%over-simplification information abstraction from data
may lead to over-simplified abstraction of original data.
%when original data have too many data points, too many observations, too large sample size, 
When original data are high-dimensional,
it adds challenges to optimizing information loss in neural network model, since there is more redundant information present. 
%original data is not fully exploited, we have big data, there is too much data to make up for the loss from not fully exploited one data, 
The main objective is not the full exploitation of original data, but to produce the highest prediction accuracy with the least amount of input data processed. 
%over-fit to one data may reduce generalization to other datasets, properly improve exploitation level of one data, helps improve generalization, find key aspects that generalize well,
Full exploitation of original data may lead to
over-fit, which reduces model generalization capability to other task data.
Well-designed exploitation methods improve data exploitation efficiency in high-dimensional data, and finds high-level generalizable representations.

%in supervised learning, back-propagation from output loss function to update whole neural network parameters, train dimension reduction unit to identify critical aspects well. 
In deep neural network, parameters in the feature extraction module are updated through back-propagation using loss gradients. 
In supervised learning, the objective can be defined as a performance-based loss containing data labels, and the dimension reduction module can be trained for the classifier to predict labels with high accuracy. 
%goal-based dimension reduction, objective-driven, reward-driven, feature extraction just to suit that particular type of goals, for that particular type of tasks
In supervised learning, extracted features are supposed to meet specific requirements in the particular type of tasks and are substantially influenced by task-specific data labels. 
For example, supervised learning may include minimization of prediction error and maximization of reward. 
These objectives contain data labels and are task-specific so that features extracted for one task cannot be generalized well to other vastly different tasks. 
%self-supervised dimension reduction, find where the sudden changes are, contrasts are, long-term and short-term trends are, which tends to be important key aspects of data, which are usually helpful later, generalize better, to other tasks with different objectives, 
Self-supervised learning objectives may include finding where abrupt changes are in time series, where sharp contrasts are in images, what long-term and short-term trends are, etc.
These critical aspects can be measured in diverse datasets by calibrating data against itself regardless of external labels. 
%for example, for data scientisit, most work lies with summary data from big data, keep data organized, average, standard deviation, these features, generalize well for all data, since it tends to be important for all tasks related to this dataset. 
For example, data scientists often conduct data cleaning and data summary such as computing averages and standard deviations. 
Data summary is generally important and can be performed in the same way for most task data.
As a result, data summary can be completed in an automated fashion by algorithms but interpretation of results still need human attention. 
In meta-learning, label-free curiosity-driven exploration that maximizes the distance between explored spots and proposed spots to explore in the future can be generalized to more tasks.

%Methods for higher level of exploitation of original data
%
%1.sufficient and complete statistics of original data, construct such mappings (rolling local or global) from original data to embeddings, then from embeddings to output
In statistical learning, a perfect embedding model should contain complete and sufficient statistics of original data. 
Learned features precisely summarize all such components of original data that are related to objective optimization. 
However, sufficient and complete statistics can be derived only for specific families of data distributions. 
%
%2.self-supervised learning,
In self-supervised learning,
%data augmentation: map original data into several versions using pure imagination, or combination with previous data experiences, mirror data
different data augmentation techniques can be applied to convert the original data into augmented versions using pure 'imagination'. 
Pre-trained model can be integrated with current task to produce more augmented versions of original data. 
%look at data from several aspects, train using these datasets, improve level of exploitation of original data
Training with different augmented data improves the exploitation level of original data and extracts more generalizable features. 
%pseudo labels for unlabelled data, through unsupervised learning, use these pseudo labels as data augmentation
Pseudo labels for unlabeled data can be computed through unsupervised clustering or pre-trained deep models from other similar tasks. 
Pseudo label is also a data augmentation technique and can be used to perform supervised learning on unlabeled data.

Nodes (activations) and weights (linear combinations of former information) constitute a layer-wise neural network model, which is used for dimension reduction in deep learning. 
If not layer-wise, neural networks may be viewed as directed graphs, which allow information to flow more flexibly through links.
Under ideal assumptions, a global optimum of information loss through any specified model structure can be attained. 
At the global optimum, model architecture is the most efficient and dimension reduction is also optimized using backpropagation of loss gradients.
Correspondingly, extracted features at the global optimum are generalizable to other tasks. 
Optimized dimension reduction in layer-wise neural network model should be closer to a global optimum when network model imposes less restrictions upon information flow. 
%
%3.information retrieval mechanisms
%
%
%various ways to account for information loss in sharp dimension reduction
In the history of neural network, several methods have been proposed to account for dimension reduction in feature extraction module.
%1. self-supervised, use pseudo data labels to supervise learning, pseudo labels are estimated using original data
%2. self-improvement, guided by pretext task pre-trained model, 
%resnet 
ResNet \cite{He2016a}
%highway network 
and highway network \cite{Srivastava2015b}
both use information retrieval mechanisms to compensate for layer-wise information loss in dimension reduction module.
ResNet uses identity mappings rather than multiplicative gate functions in highway network and achieves superior performance.  
%re-exploit original data that have not been fully exploited, to improve exploitation of original data	
By re-exploiting original data and processed information from previous layers, extracted features are more generalizable. 
%degree of dimension reduction
%information loss
%for neural network
%for statistical learning: summary/complete statistics
Information retrieval is helpful especially in the case of severe dimension reduction.

\subsubsection{Methodology of Self-Supervised Learning}
\label{Methodology}

%label-free data, more general, combine data from various sources, helps multi-modal learning
Most collected datasets do not contain human labels or annotations which are expensive to obtain. 
It is generally easier to combine unlabeled data from multiple sources or modalities for joint modelling in self-supervised learning.
%generalizable features 
Self-supervised learning primarily considers using unlabeled data to construct generalizable representations.
Generalizable features extracted using the pre-trained deep model can be applied in downstream tasks. 
Pre-training tasks bear substantial similarity to downstream tasks but are not identical to them.
Pre-trained models take advantage of higher-level aggregation and diversity in pre-training tasks to learn transferable features for downstream tasks.
Similar in meta-learning, meta-learner aggregates model training experiences to learn how to generalize and then guide self-updates of base learners to solve unseen tasks efficiently. 
Meta-learner in meta-learning performs higher-level optimization to learn how to generalize base learner, and pre-training in self-supervised learning extracts high-level generalizable features that can be applied to downstream tasks. 

%review paper of self-supervised learning
In \cite{Liu2020}, popular methods in self-supervised learning
%usually summarized to be in the following three categories
are summarized to be in three categories: 
%generative learning
generative learning,
%contrastive learning
contrastive learning,
%mixture of generative and contrastive learning
and a combination of generative and contrastive learning.
In the broad sense, self-supervised learning utilizes self-supervision to train a general model without resort to data labels. 
Generative learning simulates pseudo labels for unlabeled data from similar tasks, or performs data augmentation using rotation, colorization, crop, flip, etc. 
Contrastive learning calibrates data against itself to identify generalizable features. For example, contrastive learning minimizes the distance between augmentations of the same image and maximizes the distance between different classes of images.
Generative and contrastive learning can also be integrated in order to derive more generalizable representations of high-dimensional data. 
%in self-supervised learning, related information from other modalities can be used to boost current task solution
In self-supervised learning, related data from other modalities can be used to boost performance in current task. First, processed features from related data can be added as features to the current task. Second, data from other modalities can be used to infer unknown critical quantities in current task.

%generative: simulate data using related information from external sources
Generative scheme is also widely applied in meta-learning to simulate data that is 'indistinguishable' from real data. 
Simulated data is used as augmentation to the original data and contributes to improved performance. 
%contrastive: maximize alignment with related task solutions, semi-supervised learning
On the other hand, contrastive loss functions are label-free and optimized to learn high-level generalizable representations. 
Contrastive scheme devises label-free loss functions to maximize the alignment of predictions from the most related tasks.
Contrastive self-supervised learning is applied to train measures such that augmented images from the same original image are assigned the same label.
Contrastive objectives can also be devised through the maximization of mutual information between input data and output representations. 
% - similar to meta-learning, refer to most similar task training experiences to solve the current task
In meta-learning, label-free curiosity-based loss functions are utilized to encourage the exploration of novel spots. 
Both curiosity-driven objectives in meta-learning and contrastive loss functions in self-supervised learning are label-free and can be applied to extract higher-level generalizable features. 
%In addition, solutions to similar tasks are assumed to be close in distance so that referring to previous training expriences can be helpful in boosting current performance.
% - some require existence of memory bank 
Memory modules can also be applied in contrastive self-supervised learning to save benchmark information. 

%pre-trained large deep models, adapt, borrow pre-trained models with high accuracy, 
In self-supervised learning, pre-training is focused upon extracting more generalizable features and tackling more challenging tasks. 
Pre-training uses large deep models to learn higher-level representations.
% - similar to meta-learning, black-box adaptation line of research
In meta-learning, pre-trained deep models are adapted directly to solve unseen tasks efficiently. 
By modifying only a small proportion of parameters in pre-trained deep model, unseen tasks can be solved with high accuracy.
%generative-contrastive framework is adversarial
%maximize wrt D, minimize wrt p, mini-max game optimization, 
Generative-contrastive architecture combines a generative scheme and a contrastive scheme in an adversarial manner, where the generative scheme generates samples for the contrastive scheme to distinguish. 
In the mini-max optimization, both generative and contrastive schemes are trained leading to a more generalizable self-supervised learner. 
% - predictability minimization in meta-learning
In meta-learning, similar adversarial framework can be constructed where encoded features are inputs for the contrastive scheme to distinguish rather than generated samples. 
When encoded high-level representations truly reflect the shared similarity between different samples in the same class, it is easy to distinguish with high precision based upon these features. 

Self-supervised learning performs dimension reduction using self-supervision to formulate more generalizable features that can be further applied to downstream tasks. 
Generative self-supervision compensates for the information loss in dimension reduction module by re-visiting the original data from several different perspectives. 
Although dimension reduction remains a challenging issue, current AI methods manage to solve a wide range of tasks with sufficiently high accuracy. 
With generalizable features, we can use task training experiences to accelerate the solution of other similar tasks. 
With growing amount of diverse task data, training deep models from scratch appears less attractive and improving efficiency emerges as the new trend for AI-based data science.
The improvement brought by generalizable features is expected to complete the final step from theory to practice and to help AI applications meet the safety requirements in engineering. 

\subsection{Meta-Learning}
\label{meta-learning}

Meta-learning consists of two levels: base learner that solves task-specific issues, and meta-learner that integrates previous model training experiences to provide initialized base learner for unseen tasks. 
Meta-learning saves previous task solutions in memory module and analyzes the relation between task and solution. 
For an unseen task, meta-learning may find the most similar trained task, the solution of which can be directly applied or efficiently updated to solve the new task.  
Otherwise, meta-learning extrapolates the relation between task and solution to propose a good initial model for the new task. 
Currently, meta-learning is widely applied to design general AI algorithms and to improve automation and efficiency in robotics. 
Meta-learning serves a variety of goals: improve generalization capability of deep models, 
adapt deep models to solve vastly different tasks, 
approach diverse tasks efficiently and realize general AI, etc. 
Self-supervised learning and meta-learning both contribute to improving model generalization capability. 

\subsubsection{Overview}

\begin{figure}[htpb]
\centering
\includegraphics[width=0.57\linewidth]{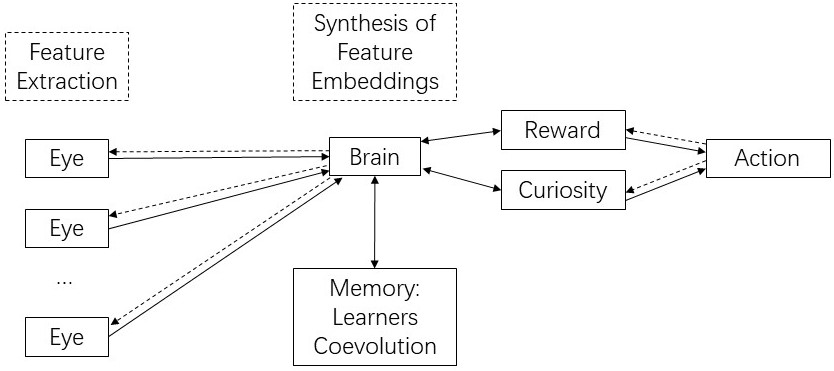}
\caption{An Example of Meta-Learning Process.}
\label{p1}
\end{figure}

Figure \ref{p1} provides an overview of processes in meta-learning framework \cite{Peng2020,Peng2021}, where feature extraction is first applied to original task data, then brain synthesizes feature embeddings to train learners. Both performance-based and curiosity-based objectives are considered where learners are also allowed to coevolve. Coevolution between learners increases the overall training efficiency of all learners and curiosity-based objectives help avoid local optimum. 

%overall structural picture of meta-learning
Learners are saved in memory modules and are allowed to coevolve. Brain represents meta-learner, which aggregates prior model training experiences to guide the training of current base learner. 
With pre-training in self-supervised learning, estimated generalizable features can be applied in feature extraction modules.
Brain communicates with memory to save and retrieve learners. Brain trains learners and feature extraction modules based upon reward and curiosity objectives. Reward maximization identifies the optimal strategy of human actions to get maximal profits. Curiosity objectives encourage algorithms to explore novel spots, to behave in different manners, to acquire new skill sets, etc.  
For vastly different tasks with different data structures, feature extraction modules are also vastly different. 
Thus, it is challenging to identify the same dimension reduction architecture that works for all tasks. 
Less assumptions should be imposed upon the data structure when devising models for general AI purposes. 
%bifurcations and applications
%\cite{Peng2020}

\begin{figure}[htpb]
	\centering
	\includegraphics[width=0.57\linewidth]{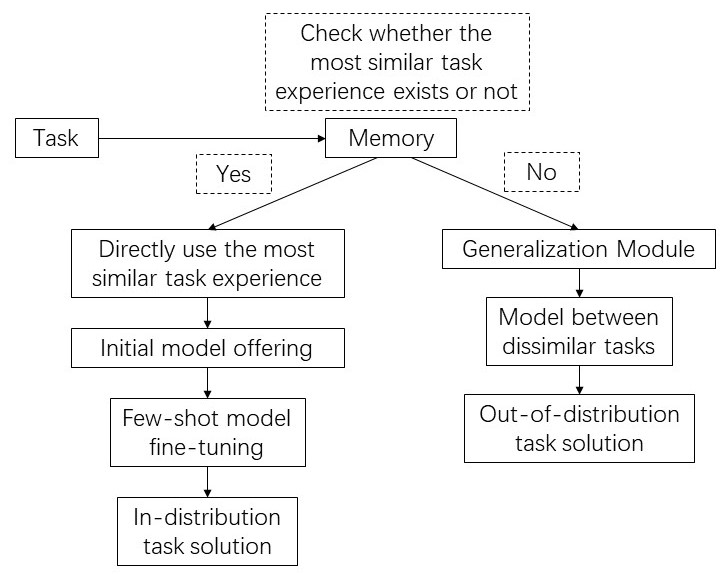}
	\caption{An Example of Meta-Learning Procedure.}
	\label{p2}
\end{figure}

Figure \ref{p2} describes a meta-learning framework with an example, where an unseen task is compared with previously trained tasks in memory. 
On one hand, in-distribution tasks are similar to trained tasks and
there exists sufficiently similar prior task experiences that can be applied to solve in-distribution tasks with only few iterations of adaptation. 
Trained models are used to offer initial models then few-shot fine-tuning is utilized to update models. 
On the other hand, out-of-distribution tasks are vastly different from trained tasks and
a generalization module is developed to extrapolate the relation between task characteristics and corresponding solutions. 
A model between dissimilar tasks measures how different tasks are and guides generalization from the aggregation of trained models to a novel task solution. 
Since the extrapolation of estimated patterns is more uncertain, generalization to vastly different out-of-distribution tasks is much more challenging.

Self-supervision in meta-learning takes several forms. 
First, feature extraction modules (eyes) perform data augmentation by varying insignificant features of original data or by simulating data that are indistinguishable from real data. 
Data augmentation guides the auxiliary training based upon original data. 
Augmented data act as self-supervision upon further training processes. 
Second, previous learners trained on similar tasks can be used as initial models.
Generated tasks are the simplest ones that trained models cannot solve.
Generated tasks are based upon trained tasks and can also be viewed as self-supervision on auxiliary training.
Efficient improvement of trained models in generalization to unexplored domains is guided by task generation processes.
Third, different learner modules coevolve and can be regarded as self-supervision upon each other. 
Regularization upon the model is constructed using data itself or other learner modules and self-supervision can be utilized in the same way as external supervision. 
Self-supervision allows efficient exploitation of the original data from more perspectives and extracts more useful features from the same data.  

\subsubsection{Connection with Self-Supervised Learning}

%intuition, we are self-driven, working hard for success, fame and money
Intuitively, we are inner-driven working diligently for reward such as success, fame and money.
%we challenge ourselves to make self-improvement
We are bored with doing repeated same tasks and  continuously challenge ourselves with more difficult ones to make self-improvement. 
Usually we work more diligently out of reward-type motivation. 
When we go to school, scores and achievements are human-made rewards that we seek to maximize. 
Likewise, out of self-motivation, humans can devise a self-score that provides reward-type motivation for themselves, and the self-score can be seen as self-supervision. 
%we have external motivation - scores and achievements, they are our labels, supervised learning, we want higher scores, exams, grades
Labels constitute our external motivation and we maximize prediction accuracy in supervised learning. 
%we have internal motivation - inner-desire, curiosity, unintentional knowledge accumulation, they are not reward-driven, purely exploration of the world, 
We have internal motivation such as inner-desire out of curiosity which leads to unintentional knowledge accumulation. 
%experience-based, from prior experience, what might be useful for future, what is different from memory items, save what can be useful for future problem solving, in memory module, 
Our exploration of the world assesses the importance of information based upon prior experience. 
Items that may be useful for future applications and novel items that are sharply different from current memory items are saved in memory module. 
%knowledge accumulation phase to improve human capability, trainning, without explicitly assigned rewards, which can be very useful later in jobs, enlarging skill sets
Knowledge accumulation phase is driven by inner-desire and curiosity and improves human capability of handling more difficult tasks. 

%illustration from meta-learning to self-supervised learning
In meta-learning, a particular form of self-supervision is task generation in powerplay
\cite{Schmidhuber2013}. 
%powerplay - coevolution between task and solver
Powerplay considers task-solver pairs and the coevolution between tasks and corresponding solutions.
% - self-generated tasks, more and more complex
Task generation mechanism produces increasingly complex tasks that are the simplest unsolvable ones for the current learner. 
% - inner-driven desire for challenge to conduct self-improvement
It simulates self inner-desire to train on more difficult tasks and to learn new skill sets for self-improvement. 
%connected with generative self-supervised learning
Generative self-supervised learning methods perform data augmentation on each task, and
augmented task data can be viewed as generated new tasks for auxiliary training. 
Powerplay uses the criterion that generated tasks should be the simplest unsolvable task for the current learner, but generative self-supervised learning does not impose such constraints upon data augmentation. 
Except for the difference in data generation mechanisms, data augmentation and task generation are both approaches to simulate more data, conduct auxiliary training and improve model performance. 
% - self-supervised self-improvement, generative desire, key to making innovation, where innovation = generalization capability
The ability to make innovations in an unseen task is closely associated with model generalization capability. 
With either data augmentation or task generation in place, trained models can solve a wider range of tasks after considering more data variations. 

%bayesian program learning
In meta-learning, another form of self-supervision is data augmentation using bayesian program learning
\cite{Lake2015a}.
%connected with generative self-supervised learning
%generative scheme, for simulated data that is indistinguishable from real data using human eye check		
Bayesian program learning estimates generative schemes and simulates new data that are indistinguishable from real data.
Simulated data and real data are jointly applied in model training. 
%simulate data from original data, imagination, combination of current data with prior data, create pseudo data that look real, 
Data augmentation in bayesian program learning is a combination of current data with previously trained task models. 
Direction of model generalization can be considered by simulating data similar to unseen tasks in the target domains. 
Required model adaptation is conducted gradually by training on carefully designed task generation mechanisms. 
%can be viewed as data augmentation from self-supervised learning
Data augmentation scheme can be viewed as self-supervision in self-supervised learning and contained in meta-learner of meta-learning. 
Meta-learner synthesizes prior model experiences and points out the direction of model adaptation for an unseen task as a form of data augmentation or task generation schemes. 
After auxiliary training devised by meta-learner, deep models reach the target generalization domain. 
%encoder-decoder paradigm
Encoder-decoder paradigm
\cite{Kodirov2017} is widely applied both in meta-learning and self-supervised learning. 
%connected with generative self-supervised learning
%encoder: from original data to embeddings
Encoder model is a dimension reduction module that maps original data to feature embeddings.
%decoder: from embeddings to more simulated data
Decoder model is a mapping from feature embeddings to simulated data. 
Encoder-decoder paradigm is utilized for task generation both in meta-learning and generative self-supervised learning. 
%no external additional information is used in this process, it is purely based upon original data
No external label is required in this process and it is purely based upon original data. 

Furthermore, 
%meta-learning using curiosity-based objective
meta-learning uses performance-based evaluations combined with curiosity-based objectives.
%objective free of performance-based evaluation
%curiosity-based objective free of labels
Curiosity-based objectives are formulated without external labels
\cite{Alet2020}. 
%connected with contrastive self-supervised learning
Likewise, in self-supervised learning, contrastive loss functions are defined using unlabeled data. 
%define objective free of observed labels, observed labels used to construct performance-based evaluations and objectives
%free of labels = free of performance-based criterion
In general, generalizable models contain dimension reduction modules that produce generalizable features.
But using labels hurts generalization capability of deep models, since for vastly different tasks, labels are also dramatically different. 
In supervised learning, observed labels are used to construct performance-based evaluations which lead to less generalizable models.
%however, we know that combination of performance-based criterion and curiosity-based criterion is most efficient
In meta-learning, it is observed that combination of performance-based criterion and curiosity-based criterion is the most efficient.
%likewise, combination of generative and contrastive self-supervised learning is most efficient
Similarly, in self-supervised learning, a combination of generative and contrastive schemes is also the most efficient. 
%objective under self-supervised learning scheme
In self-supervised learning, contrastive loss objectives can be devised to learn more transferable high-level features.
%search for a better novelty metric in an automated way \cite{Alet2020}
In meta-learning, novelty metrics can be optimized in an automated fashion such that exploration is more efficient \cite{Alet2020}. 
%curiosity-based novelty metric is often based upon tacit knowledge within task, based upon human interpretation of task information
Incorporating human interpretation of task information can extract tacit knowledge within tasks which contributes to obtaining more generalizable models. 
However, tacit knowledge is ad-hoc and cannot be programmed into automated algorithms easily. 
%automated definition of novelty metric may contribute to improved overall performance

This article discusses the connections between meta-learning and self-supervised learning. It is organized as follows. Section \ref{PretextTaskandDownstreamTask} reviews pretext tasks and downstream tasks and points out the associations with meta-learners and base learners.
Section \ref{ContrastiveScheme} surveys contrastive schemes in self-supervised learning and the extraction of high-level representations based upon mutual information maximization. 
Section \ref{GenerativeScheme} briefly summarizes generative schemes in self-supervised learning.
Section \ref{Self-PlayandGeneralAI} conveys the connections between meta-learning, self-supervised learning and general AI. 
Section \ref{ConclusionandDiscussion} concludes this paper with further discussions on the integration between meta-learning and self-supervised learning.

\section{Pretext Task and Downstream Task}
\label{PretextTaskandDownstreamTask}

\begin{figure}[htpb]
	\centering
	\includegraphics[width=0.8\linewidth]{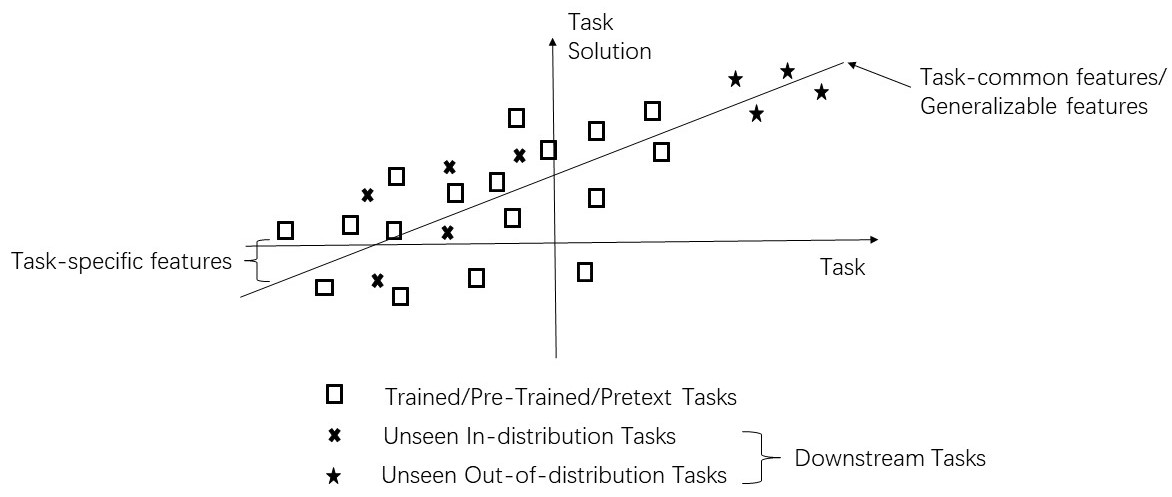}
	\caption{An Example to Show the Relation between Pre-Training and Downstream Tasks.}
	\label{p3}
\end{figure}

Figure \ref{p3} presents an example of the relation between pre-training (pretext) tasks and downstream tasks. It is shown in a 2D map for brevity. 
The 2D map depicts tasks and corresponding solutions where horizontal axis is task and vertical axis is solution. 
Using pretext tasks, pre-training model identifies a trend in the relation between tasks and solutions, where this trend represents generalizable features and the differences between pretext tasks and this trend represent task-specific features. 
Generalizable features can be transfered to other tasks including in-distribution and out-of-distribution ones. 
Out-of-distribution tasks are more different from pretext ones so that the extrapolation of pre-trained generalizable model is required and extrapolation leads to higher uncertainty in inference.

%requirements on pretext task
In general, pretext and downstream tasks are in the same category such as image or language processing.
%incorporate some intelligence and automation
The primary goal of pretext task is to identify generalizable high-level representations. 
Pretext tasks are more complex and require a higher level of intelligence and automation to solve. 
%meta-learning, pretrain on segmentation task, solve downstream image classification task,
For example, in language processing tasks, sentence ordering prediction can be used as pretext tasks. 
In image processing tasks, image segmentation can be used as pretext tasks and image classification is used as downstream tasks. 
%adaptable, freeze pretext pre-trained feature extraction module
After pre-training on pretext tasks, extracted higher-level features are transfered to solve downstream tasks. 
%pretext tasks 
%downstream tasks may be vastly different under combo of meta-learning + self-supervised learning
As features extracted in pre-training are more generalizable, corresponding downstream tasks are allowed to be more different from pretext ones. 
If learned features are not sufficiently transferable, downstream tasks are not allowed to be much different from pretext ones. 

%solving pretext task learns all techniques used to solve downstream tasks, 
After pre-training on pretext tasks, models acquire a complete set of generalizable skills that are required to solve downstream tasks.
%more complex than downstream tasks
%learn transferable features about all these tasks from pretext task
It is similar to the architecture of meta-learner and base learner in meta-learning, where meta-learner analyzes previously trained models and guides the adaptation of base learner to solve unseen tasks. 
%choose appropriate pretext task is important
%what is best tuned for pretext task should generalize well to all other downstream tasks
Choosing the appropriate pretext task is critical for high-level feature extraction since these features should generalize well to downstream tasks.

%insert a figure of linear model extraction

%for example, after model selection (embedding model), we find a model of selected right features, linear model is correct model
Here we take a linear model as an example to illustrate the relation between generalizable features from pre-training and task-specific features. 
On pretext tasks, after dimension reduction, selected features are represented by $X$ and the linear model is 
\[
y=X\beta+\epsilon,
\]
where $y$ is the pseudo label, $\beta$ is the unknown parameter to be estimated, and $\epsilon$ is the random error.
%y=xb+e
Denote projection matrix as 
$P=X(X^TX)^{-1}X^T$.
%column space is P projection matrix
Estimated linear trend in this model is
\[
\hat{y}=Py,
\]
where linear trend $Py$ is the generalizable representations from pretext tasks and it can be applied to solve downstream tasks. 
%null space is I-P
%trend is PX, this is task-common trend
%task-specific variation is I-P
On the other hand, task-specific features are 
\[
\hat{\epsilon}=y-\hat{y}=(I-P)y.
\]
%we can see that task-common and task-specific features are orthogonal to each other
In this example,
we can see that generalizable features $Py$ and task-specific features $(I-P)y$ are orthogonal to each other. 
Generalizable features $Py$ are in the column space of $X$ and task-specific features $(I-P)y$ are in the null space of $X$. 
Generalizable features $Py$ are higher-level representations from the linear model that can provide inference on downstream tasks. 
%model selection using all task data will produce features
For linear model tasks, first, model selection is conducted using all task data to select truly relevant features. 
%based upon these features, using projection, task-common features can be constructed, and task-specific features can also be identified
Second, based upon the selected features, generalizable representations of shared similarity between all tasks can be constructed using $Py$, and task-specific features are computed as $(I-P)y$.
%only trend can be extrapolated to out-of-distribution (out-of-sample) spot
Only linear trend $Py$ can be extrapolated to out-of-distribution tasks. 
%task-specific variations cannot be extrapolated
Task-specific features $(I-P)y$ cannot be extrapolated.
In unsupervised pre-training, it is critical to isolate transferable features from task-specific ones for better generalization to downstream tasks. 
%
%features = context-dependent features (downstream tasks) + context-independent features (pretext tasks)
%Features can be classified into context-dependent features specific for tasks and context-independent ones from pre-training.
%features = task-specific features (downstream tasks) + task-common features (pretext tasks)
%Features can also be divided into task-specific features in downstream tasks or base learners and high-level features in pretext tasks or meta-learner.
%features = task-specific features (base learner) + task-common features (meta-learner)
%
%feature decomposition, projection,
%regression, trend is task-common features, variation is task-specific
Features can be decomposed into projections that represent high-level features and task-specific features respectively. 
%regression over all tasks to get task-common structure, should be simpler than task-specific mechanisms, generalizable
Reregression over all tasks produces high-level representations that reflect the common properties of all tasks and can be generalized to other tasks. 
%individual features

%\cite{Bengio2011}
%deep learning of representations for unsupervised and transfer learning
\cite{Bengio2011} provides an insightful analysis as to learning representations in unsupervised and transfer learning. 
%features at multiple levels: higher-level learned feature \cite{Le2013a}, lower-level feature
Features are assumed to be at multiple levels: higher-level features \cite{Le2013a} that describe the similarity shared by diverse tasks, and lower-level features that are more task-specific and cannot generalize to other tasks.
%unsupervised pre-training of representations help learn higher-level generalizable feature
It is concluded that unsupervised pre-training can extract higher-level transferable features. 
%more abstract features
%similar to train meta-learner from pretext tasks
Pre-trained models in transfer learning and meta-learning are used for adaptation into models for unseen tasks. 
Pre-training with unsupervised learning extracts high-level generalizable features for downstream tasks. 
%meta-learner obtains task-common features that are shared similarity among all tasks
Meta-learner guides the generalization of base learner to other tasks by providing a good initial model from integrating previous training experiences. 
%pre-trained meta-learner in an unsupervised fashion
Meta-learner can be trained in supervised fashion by maximizing the prediction accuracy on validation data for all tasks. 
Meta-learner is not trained by unsupervised learning since performance-based evaluations should be used to measure model generalization capability.
%label-free data improves generalization capability (adaptivity) of learned features
Pre-training with pretext tasks uses label-free data to extract higher-level features that can be transferred to downstream tasks. 
With the same ultimate goals, 
meta-learner and pre-training with pretext tasks can be applied in the same algorithm to improve model generalization capability. 

A popular class of unsupervised pre-training model is BERT (Bidirectional Encoder Representations from Transformers)
\cite{Devlin2019a}. 
%BERT
%pre-training of deep bidirectional transformers for language understanding
%bidirectional encoder representations from transformers
%pre-train deep bidirectional representations from unlabeled text by jointly conditioning upon left and right languange context
%language model pre-training
BERT is widely applied in language processing to extract transferable higher-level representations.
%predict sentence coherece, ordering
BERT computes generalizable features for predicting sentence ordering by jointly considering left and right sentence contexts. 
%pre-training to learn generalizable higher-level features
%fine-tune all pre-trained models for adaptation
%pre-train from unlabeled text and fine-tune for a supervised downstream task
BERT is pre-trained using unlabeled texts to extract high-level features and then fine-tuned with supervised learning for adaptation to downstream tasks. 
In self-supervised learning, meta-learner can also be applied to guide the adaptation of pre-trained model to downstream tasks. 
%pre-train+fine-tune
%only pre-trained features are transfered to downstream tasks
%fine-tune bert
%moderate data augmentation
Data augmentation is also performed to enhance exploitation of original data and to improve generalization capability of deep models. 
%similar to transfer learning and meta-learning
%pre-train+adaptation
Pre-training plus model adaptation is also a typical algorithmic architecture in transfer learning and meta-learning.
%in meta-learning,
%fine-tune is guided by meta-learner
%pre-train separates task-common and task-specific features
%
Another unsupervised pre-training model is ALBERT (A Lite BERT)
\cite{Lan2019}.
%ALBERT
%a lite bert for self-supervised learning of language representations
%self-supervised loss function
%pre-trained language representations
ALBERT has the same function as BERT, i.e. to extract higher-level features pertaining to sentence ordering prediction in language processing. 
%model distillation, apply to downstream tasks
Compared to BERT, the difference is that
ALBERT conducts model distillation and then apply model to downstream tasks.
%pretraining loss and pretraining objectives
%include sentence coherence, infer ordering of sentence, words in NLP, as pretext tasks to learn generalizable task features
Sentence ordering prediction is used as the pretext task upon which generalizable higher-level representations are learned. 
%feature extraction = encoder = embedding function
%pretext pre-trained encoder
%transformer encoder

\section{Contrastive Scheme}
\label{ContrastiveScheme}

Contrastive schemes devise label-free contrastive loss functions based upon which high-level representations are extracted. 
%\cite{Oord2018}
%representation learning with contrastive predictive coding
CPC (Contrastive Predictive Coding) \cite{Oord2018}
is one of the important methods in contrastive scheme. 
CPC is originally designed in the task of time series prediction to consider long-range dependence problem. 
Later maximization of mutual information in CPC is applied to construct contrastive loss functions in self-supervised learning tasks.  
%universal unsupervised learning approach
%extract useful representations from high-dimensional data
%use a probabilistic contrastive loss
CPC uses a probabilistic contrastive loss to extract generalizable features from high-dimensional data through unsupervised learning. 
%high-level representation
%features useful to transcribe human speech may be less suited for speaker identification or music genre prediction, image captioning, image denoising
%label-free
%unsupervised learning is important towards robust and generic representation learning
Unsupervised learning uses unlabeled data and is less influenced by vastly different labels of vastly different tasks in learning generalizable features. 
%CPC learns representations that encode underlying shared information between different parts of high-dimensional signal
CPC designs contrastive objectives to be the shared information between different components of high-dimensional data. 
For example, CPC maximizes the mutual information between observations in a long time series dataset and extracts generalizable features. 
%unimodal loss function such as mean-squared error and cross-entropy loss are not very useful
%generative model minimizes reconstruction error
%in an adversarial manner
%high-level latent variables such as class labels contrain much less information
Mutual information between data $x$ and context $c$ is defined as
\[
I(x,c)=\sum_{x,c}\left[p(x,c)\log\frac{p(x|c)}{p(x)}\right],
\]
where $p(x,c)$ is the joint distribution of data and context, $p(x|c)$ is the conditional distribution of data given context, and $p(x)$ is the marginal distribution of data. 
Mutual information measures the degree of dependence between $x$ and $c$ which is the difference between $p(x,c)$ and $p(x)p(c)$.
Maximization of mutual information identifies the most probable context $c$ for data $x$. 
We know that
\[
p(x,c)=p(x|c)p(c).\]
Then log ratio is
\[
\log\frac{p(x,c)}{p(x)p(c)}=\log\frac{p(x|c)}{p(x)}.\]
Integral of this difference is mutual information
\[
\int_{x,c}p(x,c)\log\frac{p(x,c)}{p(x)p(c)}dxdc=I(x,c).
\]
Mutual information between data $x$ and context $c$ describes the amount of dependence between them.
We can see that mutual information is label-free and the maximization of mutual information leads to the most probable context $c$ for original data $x$. 
We may maximize mutual information in the following ways to construct contrastive loss functions:
(1) between original data and latent context variables,
(2) between encoded features of different components in original data,
(3) between pretext task data and downstream task data, etc.
%maximize mutual information between original signal and context latent representation
%maximize mutual information between encoded representations
%maximize mutual information between pretext and downstream tasks
Maximization of mutual information is a powerful way to construct contrastive loss functions in contrastive self-supervised learning.
By maximizing mutual information, components with the highest similarity to each other are aligned.   
%downstream tasks require pooled representation over all locations
%extract slow features, maximize mutual information of observations over long time horizon
%define prediction task from related observations, extract useful features
%contrastive predictive coding (CPC), compact latent representations to encode predictions over future observations
In CPC, an embedding function is applied to original data: $z_t=f(x_t)$. 
Latent context variables are based upon features of observed data: $c_t=g(z_1,\cdots,z_t)$.
%learn abstract representations in an unsupervised fashion
Estimation of mutual information is performed in the following steps. 
First, construct a random sample of size $N$ with $1$ sample generated from $p(x_{t}|c_t)$ and $N-1$ samples from $p(x_{t})$.
The probability that the $k$th sample onward $x_{t+k}$ is from distribution $p(x_{t+k}|c_t)$ is denoted by $f(x_{t+k},c_t)$ and $f(x_{t+k},c_t)$ is proportional to the ratio $p(x_{t+k}|c_t)/p(x_{t+k})$.
To estimate $f(x_{t+k},c_t)$, minimize loss function:
\[
L=-\mathbb{E}_x\left[\log\frac{f(x_{t+k},c_t)}{\sum_{x_j}f(x_j,c_t)}\right].
\]
We know that $f(x_{t+k},c_t)$ is proportional to the ratio $p(x_{t+k}|c_t)/p(x_{t+k})$ and thus is proportional to mutual information $I(x,c)$. We find the maximal value of $f(x_{t+k},c_t)$ that corresponds to the maximal value of mutual information. 
Here in the example of time series prediction, estimated long-range dependence structure is the generalizable feature. 
Contrastive loss objective is based upon the mutual information between different components in the same high-dimensional data. 
Similar design of contrastive loss functions can be composed in self-supervised learning to extract generalizable features of interest.
In general, mutual information can be defined, estimated and maximized in self-supervised tasks. 
Maximization of mutual information is an important tool to define powerful contrastive loss functions. 

DGI (Deep Graph Infomax)
\cite{Velickovic2019} handles graph-structured task data and applies contrastive self-supervised learning to learn high-level node representations. 
%deep graph infomax DGI
%learn node representations within graph-structured data in an unsupervised manner
%maximize mutual information between patch representations and corresponding high-level summaries of graphs
Similar to CPC, mutual information between different patches within the same high-dimensional graph-structured data can be maximized to construct contrastive loss functions. 
%train an encoder to be contrastive between representations that capture statistical dependencies of interest and those that do not
Like in CPC, maximization of mutual information between local patches can be applied to identify critical dependencies that can be generalized as high-level features. 
%train an encoder model by maximizing mutual information between a high-level global representation and local parts of input
DGI maximizes the mutual information between generalizable global features and local components in high-dimensional data. 
%leverage local mutual information maximization across graph's patch representations to obtain node embeddings
%contrast global and local representations on graphs
Global and local features are contrasted over all patches to learn generalizable high-level representations. 
%maximize mutual information between input and output
%between local neighborhood and high-level features
%potential of methods based on local mutual information maximization in the inductive node classification domain
Since the number of patches and nodes is large, the maximization of mutual information over all local patches may be accelerated using more efficient search algorithms. 
%apply mutual information maximization to graph-structured inputs
%train an encoder network so that nodes that are close in the input graph are also close in the representation space
Furthermore, an encoder model is trained such that representations of close nodes in the original data are also close after dimension reduction. 
%deep infomax (DIM) learn representations of high-dimensional data
%deep graph infomax (DGI)
%scoring function
%conditional density
%learn unsupervised representations on graph-structured data

SimCLR (Simple framework for Contrastive Learning of visual Representations)
\cite{Chen2020} is another method for contrastive self-supervised learning. 
SimCLR does not use mutual information to formulate contrastive loss objectives. 
%SimCLR
%a simple framework for contrastive learning of visual representations
%without memory bank
%different data augmentation techniques applied to the same image
Several different data augmentation techniques are applied to the same image, such as rotation, colorization, flips, etc. 
%contrastive loss function
The contrastive loss function is constructed by 
%compare difference between images
%minimize difference of different data augmentation techniques applied to the same image
assigning all augmented versions of the same image to be in the same class. 
%maximize agreement of image embeddings
SimCLR maximizes the alignment of extracted features from all augmented versions of the same image to train an encoder model. 
Encoded representations are generalizable features that can be applied in other tasks. 
Rather than maximization of mutual information, SimCLR builds contrastive loss functions in a more intuitive way and achieves satisfying performance. 

Label-free objective functions are also widely applied in meta-learning
\cite{Lehman2011},
%in meta-learning, objective is abandoned, no data label is present, curiosity-driven approach, more general
such as curiosity-based objectives. 
In order to avoid local optimum and identify solutions closer to global optimum, curiosity-driven exploration is employed to tour the whole search space in the hope of finding a better solution. 
Curiosity-driven approaches encourage trying novel skill sets to experiment with their performances. 
%free from labels
%similar to contrastive learning
%no data label required
%more general
%less trapped at local optimum
At global optimum, trained models show the best performance with the target generalizable features extracted. 
Curiosity-driven optimization also belongs to the category of unsupervised learning since no data label is required. 
Curiosity-based methods in meta-learning appear similar to contrastive self-supervised learning, though the underlying
intuitions are different. 
%in self-supervised learning, metric learning is used to maximize distance between different classes, to minimize distance between same class
%in meta-learning, curiosity-driven approach maximizes novelty of spots to be explored, minimizes similarity with explored spots, seek novel potential
Novelty metrics which are the curiosity-based objectives to optimize in meta-learning should be devised according to the requirements of particular tasks.
Similar in contrastive self-supervised learning, contrastive loss objectives should be defined based upon task requirements. 
In meta-learning, curiosity-based criteria using unlabeled data and performance-based criteria with labeled data are combined in applications to boost performance.
Similar in self-supervised learning, contrastive schemes and generative schemes are integrated to learn high-level representations.

\section{Generative Scheme}
\label{GenerativeScheme}

%data augmentation
Generative schemes are widely applied in machine learning. 
A typical example of generative scheme is data augmentation such as rotation,
%rotation
colorization \cite{Zhang2016a},
%colorization \cite{Zhang2016a}
flip and crop,
%flip and crop
%noise infusion
noise infusion, perturbation, etc. 
Generative self-supervised learning can be applied to improve the performance on few-shot image tasks
\cite{Gidaris2019}.
%boosting few-shot visual learning with self-supervision
%different facets of the same problem
%with little or no labeled data
%low data regime
%unlabeled data
In the setting of low data and high dimension, 
generative schemes approach the same data from different perspectives and extracts more generalizable representations from original data.  
%self-supervision as an auxiliary task
In generative self-supervised learning, augmented versions of task data are used as self-supervision to train deep models. 
%learn richer and more transferable visual representations
With more variations of original data considered, trained models learn higher-level transferable features and can be better generalized to other tasks. 
%use labels from other similar datasets
For unlabeled data, self-supervised learning can estimate pseudo labels from trained models on similar task data. 
%add a self-supervised loss to training loss
A self-supervised loss function can be added to the supervised loss with pseudo labels in order to learn high-level representations.
%hallucinate additional examples from reduced amount of data
A fundamental component of a generative scheme is to 'imagine' new data from observed data and to use augmented data as self-supervision in auxiliary training. 
%annotation-free pretext task
%surrogate supervision for feature learning
%features useful for downstream tasks
%few-shot classification loss+self-supervision loss
%rotation augmented image alignment: augmentation from same images is assigned the same class

%rethinking pre-training and self-training
%data augmentation is guidance for generalization module, which is the meta-learner, should be adaptive, considered in
Data augmentation is a critical part of generative schemes, where augmented data are expected to cover unseen cases that deep models will generalize to solve in the future. 
%meta-learner, with more tasks trained, more data augmentation techniques might be added in training
Meta-learner saves previous model training experiences and provides building materials for data augmentation techniques. 
%data augmentation as control factors
%pre-training used to initialize deep models
Pre-trained models and models recommended from meta-learners can be used as initiations for deep models. 
%initialize backbone models such as deep metric models, deep feature extraction models
Self-supervised learning methods have flexible forms and architectures. 
Self-training \cite{Zoph2020} combines self-supervised learning with auxiliary supervised learning on augmented data. 
%self-training=self-supervised learning
%+additional auxiliary augmented data of the same data structure, supervised learning
%contrast additional task data against pre-training
%self-training:
Self-training contains the following steps.
%1. remove labels from imagenet
%2. train object detection model on COCO (self-supervised pre-training), learn more universal representations
(1) Supervised pre-training is conducted on labeled task data. 
%3. use object detection model to generate pseudo labels on imagenet (data augmentation)
(2) According to pre-training, pseudo labels are generated for unlabeled data. 
%4. pseudo-labeled imagenet and labeled COCO combined to train a new model
%self-training is based upon noisy student training
%1. teacher model is trained on labeled data (COCO)
%2. teacher model generates pseudo labels on unlabeled data
%3. student model is trained to optimize loss on human labels and pseudo labels jointly
(3) Unlabeled data with pseudo labels and originally labeled data are jointly used to train a deep model. 
In self-training, self-supervision is from pseudo labels proposed from pre-training. 
The key to a generative scheme is that augmented data are 'indistinguishable' from real data so that no harmful bias arises from data augmentation processes. 

%ways to pool data together to get task-common features
To obtain higher-level representations, more task data should be pooled together to make an integrated analysis. 
%how to merge results from different data sources
For unlabeled data, it is usually easier to aggregate data from multiple sources and modalities. 
%in meta-learning, base learner results are pooled to update meta-learner (task-common)
Similar in meta-learning, training results of base learners are pooled to update meta-learner. 
Generative mechanisms are helpful especially in few-shot learning when available training data are of small sample sizes. 
Augmented data provide us with different perspectives upon the same dataset to compensate for the information loss due to dimension reduction in encoder models. 
Extracted features are more generalizable
when more variations of the same data are considered. 
%in self-supervised learning, augmentation, pseudo labels from various tasks, pooled, to solve data

\section{Self-Play and General AI}
\label{Self-PlayandGeneralAI}

%\cite{Peng2021}
%meta-learning and general AI are closely related
Meta-learning and general AI are closely related \cite{Peng2021},
%self-supervision: coevolution between learners
where coevolution modules between different learners in the same system or different components in the same learner constitute self-supervision upon each other. 
For example, objective functions of learners take inputs from each other so that learners updates are intertwined. 
The information flow between learners forms self-supervision upon related components so that they must stay tuned with each other for the system to be updated efficiently. 
%coevolution between learner and generated task
Generated challenging tasks can also be viewed as self-supervision for training more generalizable learners. 
%powerplay
%novelty-driven and performance-driven objective
Novelty-driven and performance-driven objectives explore more possibilities over the search space to avoid local optimum. 
%coevolution
%
%self-motivation
%self-supervision
In a sense, self-supervision in this realm is derived from self-motivation to seek challenging tasks in training for better self-improvement. 
%out of curiosity
Self-supervision may also stem from curiosity for novel tasks and new challenges.
%learners evolve to be increasingly complex
%meta-learning
%general AI
%utilize self-supervision to generalize deep models
%self-supervision
Self-supervision requires oneself to be aware of its own weaknesses and be capable to devise challenging tasks that are the most appropriate for its own self-improvement. 
%play against oneself
%identify weak spots, design challenging tasks, to improve upon one's weak spots

Usually in reinforcement learning tasks such as Atari 2600 games, game rules are not explicitly provided to algorithms as input data. 
Learners do not explicitly output the estimation of game rules and make self-improvement only through playing the games repeatedly and watching the videos. 
We can see that learners can play the games well after a long time of training. 
General AI algorithms can improve generalization capability of deep models and eventually build automated systems with versatile skills to conquer diverse tasks.  
%\cite{Silver2018}
%coevolution between learners
%coevolution between learner and generated task
%AlphaGo Zero
AlphaGo Zero \cite{Silver2018} is an improved version of AlphaGo and can beat the best players in chess, shogi and Go games. 
AlphaGo can only be applied to Go games and requires a long time of training to win.  
AlphaGo Zero follows a different strategy from AlphaGo, and is designed to be a general AI algorithm that can solve a wide range of tasks. 
%input explicit game rules
AlphaGo Zero takes game rules as input data and seeks self-improvement through self-play, where learners play against the best generations as challenges. 
%through self-play
%beat human best players in chess, shogi and Go games
%more general than AlphaGo

%AlphaZero
%maximize expected outcome of games
The ultimate objective of AlphaZero \cite{Silver2018} is the maximization of expected game outcomes.
%reinforcement learning from self-play games, starting from randomly initialized parameters
%updated parameters are used in subsequent games of self-play
AlphaZero trains deep reinforcement learning models with random initiations and updates parameters through self-play.
%self-play games are generated by the best player from all previous iterations
In each generation, the best learners ('players') are retained and used to generate self-play games as new challenging benchmarks to play against. 
%each search consists of a series of simulated games of self-play
%self-play games are generated by the best player from all previous iterations
%self-play games are always generated by using the latest parameters for this neural network
%time spent on generating self-play games = 300* time spent on training neural networks
The process of simulating self-play games is very time-consuming and costs about 300 times more resources than training deep network models. 
Designing the most appropriate training tasks for the current learner is critical for the most precise and efficient self-improvement upon weaknesses.
%require no domain knowledge 
%no need to consider adaptation
%general RL algorithm that learn any game
%trained within hours
AlphaZero takes much less time than AlphaGo and can be trained to solve games in hours. 
%input explicit game rules: which may take a long time for an algorithm to learn merely from input videos or demonstrations
Generative schemes simulate self-play games from previously emerged best learners and guides current learners to compete with the previous best ones for self-improvement.
Self-play games constitute self-supervision in this general AI architecture, which can also be viewed as a meta-learning framework with coevolution embedded. 
AlphaZero not only provides a general algorithm to play diverse games, but also improves model efficiency by only playing against the current best self-play games.

\section{Conclusion and Discussion}
\label{ConclusionandDiscussion}

%meta-learning combined with self-supervised learning
Meta-learning can be combined with self-supervised learning to create new techniques for better model generalization property. 
%model adaptation with label-free data
Self-supervised learning uses unlabeled data to learn transferable representations. 
%supervised model adaptation
%unsupervised model adaptation
Meta-learning usually processes labeled data and can also process unlabeled data after integration with self-supervised learning techniques.
%model adaptation guided through self-improvement
%self-improvement does not require data labels
Self-supervised learning can perform model adaptation and self-improvement under the guidance of meta-learner, which analyzes model training experiences to offer suggestions. 
%objective-driven tasks
%label-free objective
%curiosity-driven objective
%contrastive loss function
Contrastive loss functions can be optimized with the help of curiosity-driven objectives to avoid local optimum. 
%metric learning loss, ranking-based, 
%generalization-driven objective,
Task objectives can be optimized jointly with 
measures of model generalization capability to train deep models.  

%meta-learning + self-supervised learning for few-shot learning
We know that meta-learning makes a significant contribution to few-shot learning.
Self-supervised learning can also be applied to boost the performance of few-shot learning and to process few-shot unlabeled data. 
On one hand, self-supervised learning
provides data augmentation techniques, offers pseudo labels acquired from similar task training experiences, extracts generalizable features using unsupervised pre-training, and constructs contrastive loss objectives for unlabeled data. 
On the other hand, self-supervision can be learned adaptively in an automated fashion to maximize performance. Self-supervision can be combined with meta-learner to design better general AI algorithms. 
%data augmentation, pseudo labels learned
%few-shot data
%meta-learning applies pseudo labels for model adaptation to different tasks

From meta-learning and self-supervised learning, methods to improve model generalization include the following. (1) Unsupervised pre-training is applied to extract generalizable features for downstream tasks. (2) Meta-learner analyzes prior model training experiences to learn the relation between task and solution, then provides a good initial model for an unseen task. (3) A contrastive loss function can be formulated using mutual information maximization and is optimized to extract high-level representations. (4) Curiosity-based algorithm is applied to search for global optimum. 
The ultimate objectives of meta-learning and self-supervised learning are similar and methods from these two areas can be combined in the same algorithm to improve overall model generalization capability.

\section{Conscience and Consciousness}

Self-supervision can be constructed by looking at oneself from multiple perspectives to discover one's advantages and disadvantages \cite{Pitrat2009}. Current self is compared with the previous one to identify whether prior changes have led to improved performance. 
Different perspectives on oneself are generated from bootstrapping memory items \cite{Phillips2019}. 
Several copies of one self are made, where one competes with itself in coevolution \cite{Dormoy}.
Similar to data augmentation techniques, bootstrapping creates duplicates that inherit critical properties of the original system. 
At the meantime, bootstrapping is a higher-level process that contains model self-updates and self-modifications \cite{Phillips2019}.
Self-supervision here is built upon a meta-level 'brain' that directs bootstrapping and a mechanism which drives the coevolution between different selfs.
In artificial general AI systems, both interactions between different learners and interactions between learners and environment are considered. 
In Meta-Reinforcement Learning, environment produces reward to agents through interactions in-between. On the other hand, interactions between different agents are viewed as cooperation or competition, which have an indirect effect upon external reward from environment. 

Both conscience and consciousness are equipped in an artificial general AI machine \cite{Pitrat2009}. 
Consciousness refers to the incentive and motivation to 'brainstorm' viable actions for the machine to choose from. 
Conscience refers to the ability to judge how good an action is. 
In 'Self-awareness', an AGI machine compares itself with multiple bootstrapped selfs to see its own benefits and drawbacks. 
A typical AGI system keeps running for several years accumulating task experiences and constantly making self-improvements. 
In the lifetime of an AGI machine, its conscience and consciousness are continually updated adding more principles learned from generated tasks. 
In this sense, machine conscience and consciousness are constructed through self-supervision. 

In an AGI structure, a meta-trace records machine reasoning processes which make decisions on the best action to take \cite{Pitrat2009}. 
A trace saves the mechanical results which include detailed evaluations upon each action. 
Meta-traces contain meta-explanations as to why machines choose actions following the devised procedures. 
In meta-learning, a meta-learner is designed for higher-level analysis of previous task experiences.  
Several meta-levels can be stacked in the same general AI system \cite{Dormoy}, where upper levels generalize from model experiences in lower levels. 
From \cite{Pitrat2009} and \cite{Dormoy}, we can see that self-supervision and meta-learning notions are indeed combined to form artificial general intelligence systems.
We can see correspondences between "meta-learner and base learner meta-learning structure" and "meta-trace and trace conscience and consciousness architecture". Discussions are as follows \cite{Pitrat2009}.

1. Meta-trace keeps the reasons and principles based upon which we design task-specific decision models. These reasons are named meta-explanation. Meta-explanation is generalizable and shared by all task-level decision models.
Meta-trace performs the function of meta-learner by providing an initial decision model for a new task and by extracting meta-explanation knowledge from trained tasks. 

2. Trace keeps records of action evaluations from all decision models in tasks. Base learner adapts the initial model to solve current tasks and reports training results to meta-learner for further integrated analysis. 
In this sense, trace and base learner are both concentrated upon task-specific models. 

3. Meta-trace covers all principles related to model building in AGI systems. Meta-trace accounts for conscience and does not consider consciousness, since meta-trace only contains higher-level reasoning rules, also known as meta-explanation.  
Bootstrapper is also guided by principles in meta-trace, where bootstrapping creates copies of oneself and makes self-modifications.

Can machines learn emotions? \cite{Pitrat2009} Are emotions natural generalizations from solving diverse unemotional tasks? 
From the generalization mechanims in AGI systems, can emotions be created from non-emotional missions?
Generalization is an innovative process, where we learn generalizable features from prior tasks, and transfer pre-trained patterns to new tasks. 
The manner in which we update pre-trained models is also derived from previous non-emotional tasks. 
In generalization, we do not resort to ideas out of nowhere to make self-modifications. 
There is no emotional component in data collection protocols, objective functions, self-updates, or anywhere else. 
Are emotions implicitly included in the original training data which are collected upon emotional bodies such as humans?
No.
We cannot tell the difference between emotional behaviors and non-emotional behaviors by looking at data. 
The effects of emotions are versatile, unpredictable, confounded with other variables, untraceable and hard to quantify. 
In such settings, it is almost impossible (with probability close to 0) for artificial beings to acquire emotional behaviors. 

Free our imagination for one second. We want artificial beings to function as if they had human-like emotions, then we must collect data on humans to tell machines what emotions mean. 
For example, emotions may lead to sub-optimal or beyond-optimal results. In examinations, negative emotions make students score lower but positive emotions make athletes score higher. 
Emotions cause observed patterns to deviate from pre-trained patterns learned from non-emotional missions. 
As a result, to improve prediction accuracy under the influence of emotions, we learn the additional component: how emotions change learned patterns. 
Even so, "as-if emotions" in machines are not the same as human emotions, since they are derived from completely different sources. 
Machines imitate observations from humans, produce similar behaviors, but do not follow exactly the biological processes that give birth to emotions in humans. 
Hence, to this point, machines are still just imitators of observations, but are not capable of 'feeling' versatile unpredictable human emotions.

\section*{Acknowledgment}

I appreciate valuable comments from Basile Starynkevitch <basile@starynkevitch.net>. I am grateful for valuable publications from Jacques Pitrat, which describe his research work on artificial general intelligence in detail. 

\bibliographystyle{unsrt}

%\bibliography{appl2}


\end{document}